\documentclass[sigconf]{acmart}

\usepackage{graphicx}
\usepackage{booktabs}
\usepackage{multirow}
\usepackage{tabularx}
\usepackage{enumitem}
\usepackage{amsmath}
\usepackage{url}

\AtBeginDocument{%
  
}

\renewcommand\footnotetextcopyrightpermission[1]{}
\begin{document}


\title[Multimodal Detection of Higher-Order Behavioral Constructs]{Multimodal Detection of Higher-Order Behavioral Constructs: Self-Compassion in Structured Reflective Interaction}


\author{Siddhant Jain}
\orcid{0009-0005-3051-7064}
\affiliation{%
  \institution{Deutsches Forschungszentrum f{\"u}r K{\"u}nstliche Intelligenz GmbH (DFKI)}
  \city{Saarbr{\"u}cken}
  \state{Saarland}
  \country{Germany}
}
\affiliation{%
  \institution{Universit{\"a}t des Saarlandes}
  \city{Saarbr{\"u}cken}
  \state{Saarland}
  \country{Germany}
}
\email{siddhant.jain@dfki.de}

\author{Dimitra Tsovaltzi}
\orcid{0000-0002-1670-5799}
\affiliation{%
  \institution{Deutsches Forschungszentrum f{\"u}r K{\"u}nstliche Intelligenz GmbH (DFKI)}
  \city{Saarbr{\"u}cken}
  \state{Saarland}
  \country{Germany}
}
\affiliation{%
  \institution{Universit{\"a}t des Saarlandes}
  \city{Saarbr{\"u}cken}
  \state{Saarland}
  \country{Germany}
}
\email{dimitra.tsovaltzi@dfki.de}





\begin{teaserfigure}
\centering
\includegraphics[width=0.85\linewidth]{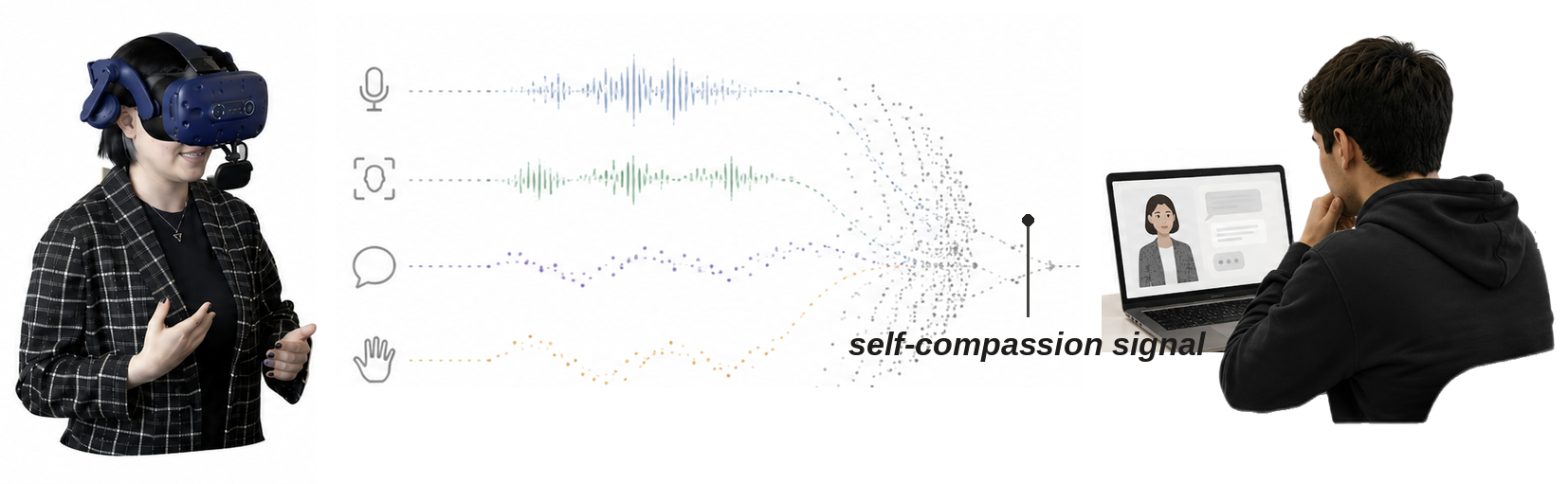}
\caption{A participant interacts with a technology-mediated training scene (left), illustrating the kind of structured reflective interaction studied in this work, alongside a related moment of self-reflection in a digital context (right).}
\label{fig:teaser}
\end{teaserfigure}

\begin{abstract}
Many of the qualities that matter most in how people learn and grow, how someone regulates their emotions, reflects on a setback, or stays aware of others during a difficult conversation, are not directly observable. They have to be inferred from how someone speaks, moves, and sounds over time, and they resist the kind of clean labeling that most machine learning pipelines are built around. We study this challenge through a case that is well grounded in psychological theory but rarely modeled computationally: self-compassion, the tendency to respond to one's own setbacks with patience rather than harsh self-criticism. We examine how it appears during structured reflective interviews in a technology-mediated training setting, where people naturally talk through socio-emotionally demanding situations. Since no existing dataset captures this kind of construct in this kind of setting, we collected and annotated 51 reflective dialog sessions using an independent, temporally overlapping annotation scheme grounded in established theory. We consolidate the underlying six-component psychological model into a three-class supervision space, balancing self-kindness and mindfulness against self-critical or overwhelmed states, and build a reproducible window-based pipeline that aligns video, audio, and text on a shared timeline. Unimodal models trained on each modality separately are compared against a simple probability-level fusion strategy, which yields modest but consistent gains over the best single modality. We close by discussing where each modality succeeds or struggles, what this suggests about how this kind of construct is actually expressed in reflective speech, and what would be needed to model it, and constructs like it, more effectively.
\end{abstract}

\maketitle
\pagestyle{plain}
\thispagestyle{plain}

\section{Introduction}
\label{sec:intro}

Many of the qualities that matter most in situated, interactive settings, reflective regulation, collaborative awareness~\cite{tsovaltzi2015}, adaptive self-evaluation, are not directly observable. They have to be inferred from behavioral signals that unfold over time and are embedded in ongoing interaction. Unlike basic emotion categories, these constructs are theory-driven and conceptually structured rather than directly measurable: their expression is distributed across language, vocal prosody, and non-verbal movement, and the expert annotations used to study them are typically sparse, temporally extended, and overlapping. Together, these properties make such constructs difficult to represent, supervise, and evaluate with standard machine learning tooling.

Self-compassion may serve as an exemplary case for studying this difficulty, since it comes with an established psychological framework~\cite{neff2003,neff2009} but has rarely been examined outside self-report questionnaires. We study it as it manifests during structured reflective interviews. Reflective practice is a central mechanism in professional development~\cite{schon1983}, but reflection can also intensify rumination or harsh self-evaluation when emotional regulation is ineffective, and lead to counterproductive over-identification experiences~\cite{nolenhoeksema2008,leary2007}. Self-compassion offers a structured account of adaptive self-regulation under exactly this kind of difficulty, and teachers and teacher trainees who report higher self-compassion show lower burnout and greater professional well-being~\cite{moe2020}. What has remained largely unexplored is how self-compassion actually shows up, moment to moment, while someone is reflecting out loud.

To study this, we draw on data collected within a technology-mediated training system designed to support socio-emotional skills in high-stakes professional interactions~\cite{chehayeb2025effectscoregulationmodelmr}. Participants first take on a professional role in a simulated conflict scenario and are then guided by a human interviewer through a structured reflection on their own recorded behavior. Because no existing multimodal corpus targets self-compassion in this kind of structured reflective setting, we collected and annotated a dedicated dataset as part of this project.

The paper makes four contributions: a transformation from theory-driven, independently annotated constructs into a modeling-compatible multi-label representation, consolidating the original six-component scheme into three classes under empirical support constraints; a reproducible multimodal preprocessing and alignment pipeline covering video, audio, and text under a shared window-based temporal backbone; a systematic unimodal evaluation of video, audio, and text under an identical supervision protocol; and an evaluation of probability-level late fusion as a controlled, interpretable multimodal integration strategy. We present these as baseline results for a task that, to our knowledge, has not previously been addressed in this combination of construct, setting, and modeling formulation. Thus, we contribute to methodologies for modeling higher-order behavioral constructs more broadly, using self-compassion as one instantiation of this problem rather than as an end in itself.
\section{Background and Related Work}
\label{sec:background}

\subsection{Self-Compassion as a Regulatory Construct}
Neff formalizes self-compassion as a multidimensional construct with three bipolar dimensions~\cite{neff2003,neff2009}. \emph{Self-Kindness} versus \emph{Self-Judgment} concerns whether one responds to personal difficulty with care or with harsh criticism. \emph{Common Humanity} versus \emph{Isolation} concerns whether suffering is recognized as a shared human experience or perceived as uniquely personal. \emph{Mindfulness} versus \emph{Over-Identification} concerns whether painful thoughts and emotions are held with balanced awareness or become overwhelming. Self-compassion is theoretically distinct from self-esteem, which is typically contingent on performance and comparison, and from self-pity, which may exaggerate personal suffering without maintaining perspective~\cite{neff2009}.

Although originally described as bipolar, empirical work suggests the positive and negative poles can co-occur and fluctuate independently within short intervals~\cite{strauss2016}, motivating a multi-label rather than mutually exclusive formulation. Reflective episodes frequently involve perceived mistakes, interpersonal conflict, or uncertainty, and when emotional regulation is insufficient, reflection can slide into rumination or harsh self-evaluation~\cite{nolenhoeksema2008,leary2007}.

\subsection{Behavioral Correlates of Self-Compassion}
Facial expression and vocal prosody are broadly understood to carry affective and regulatory information~\cite{ekman1997,scherer2003}, and the most directly relevant evidence for self-compassion specifically comes from work analyzing facial and acoustic expressions of self-compassion, self-criticism, and self-protection in emotion-focused therapy sessions~\cite{bailey2023,bailey2024}. These studies show that self-compassion-related states manifest in measurable facial and vocal behavior, but they are unimodal, use event-based rather than window-aligned supervision, which fixes a consistent temporal unit across modalities rather than variable-length labeled events, and do not address cross-modal alignment or multi-label component structure. Our work extends this line of evidence to a different setting, structured professional reflection rather than therapy, and to a multimodal, multi-label formulation that captures overlapping, co-occurring signal a single-label or single-modality design would miss. 

\subsection{Multimodal Modeling of Complex Behavioral Constructs}
Recent multimodal affective interaction modelling has combined pretrained modality-specific encoders with transformer-based fusion, including cross-modal attention~\cite{shayaninasab2024}, dense shared-private representations~\cite{deng2023}, noise-resistant training~\cite{liu2024}, and reliability-aware weighting~\cite{waligora2024}. These architectures perform well on established benchmarks, but that performance typically assumes short clips, single-label supervision, and comparatively large training corpora. Reflective dialog data looks different on all three counts: sessions are long, annotations are overlapping and multi-label, and the number of sessions is small, conditions under which high-capacity joint architectures are prone to overfitting rather than learning useful cross-modal structure. We therefore adopt a lower-capacity, modular design instead: independent unimodal encoders combined through probability-level late fusion, which avoids introducing additional trainable cross-modal parameters and keeps each modality's contribution interpretable. Detection itself is formulated as multi-label classification~\cite{zhang2014multilabel}, since self-compassion's components are theorized to co-occur rather than to be mutually exclusive~\cite{strauss2016}, which allows leveraging a rich context of interaction to model abstract constructs which otherwise remain underspecified.

On the language side, cross-modal and instruction-tuned models such as CM-BERT~\cite{yang2020cmbert} and Emotion-LLaMA~\cite{cheng2024emotionllama} demonstrate that pretrained and instruction-following language models capture affectively relevant textual signal, and linguistic markers such as pronoun use and self- versus other-focused language have been shown to index self-relating and interpersonal stance in psychotherapy transcripts~\cite{ryu2023}. Comparatively little work, however, has examined whether general-purpose or instruction-following language models capture theory-grounded, multi-component constructs like self-compassion without task-specific adaptation, which motivates the fine-tuning approach we take for the text modality.

This gap reflects a broader pattern: multimodal modeling of higher-order, theory-grounded constructs remains underexplored outside a small set of benchmark emotion categories. Self-compassion, and the structured reflective setting in which we study it, offers one concrete route into this broader problem, which is what motivated the data collection and modeling approach described in the remainder of this paper.
\section{Problem Formulation}
\label{sec:problem}

\subsection{Research Questions}
This work is guided by four questions.

\textbf{RQ1:} Can an abstract theory-driven, six-component psychological construct be consolidated into a supervision scheme that remains faithful to the underlying theory while matching the empirical support available in the collected corpus?

\textbf{RQ2:} What alignment and supervision protocol is needed to bring facial, acoustic, and linguistic signal, each sampled at a different rate and subject to different sources of noise, onto a shared temporal backbone suitable for cross-modal comparison?

\textbf{RQ3:} When modeled independently under an identical protocol, do video, audio, and text carry comparable, complementary, or redundant information about self-compassion as it is expressed during structured reflective dialog?

\textbf{RQ4:} Does a low-capacity, interpretable late fusion strategy recover meaningful complementary signal across modalities, and how large is that gain relative to the strongest single modality?

\subsection{Label Consolidation}
All six components were annotated independently as overlapping ELAN tiers. For supervised modeling, we consolidate this into a three-class space: Mindfulness (MF), Self-Kindness (SK), and Negative components (NEG), where NEG aggregates Self-Judgment and Over-Identification. Common Humanity and Isolation were excluded due to sparse temporal coverage (121s and 15s of total annotated duration across 5 and 1 videos respectively, versus 7{,}137s for MF and 3{,}056s for SK). This consolidation reflects empirical constraints in the collected corpus, not a theoretical claim that the excluded or aggregated components are equivalent; we return to this trade-off in Section~\ref{sec:limitations}.

\subsection{Component Co-Occurrence}
Prior work indicates that self-compassion components may fluctuate independently within short temporal intervals rather than behaving as mutually exclusive states~\cite{strauss2016}: a speaker may express mindful awareness while simultaneously engaging in partial self-criticism, or express self-kindness alongside over-identification. This partial independence motivates treating detection as a multi-label rather than single-label problem, allowing overlapping regulatory tendencies to be identified concurrently within the same temporal window.

\subsection{Multi-Label Formulation}
The task is formulated as multi-label classification: each unit of analysis receives independent binary labels for \{MF, SK, NEG\}, and a model produces independent sigmoid-linked probability estimates for each class without cross-class normalization.
\section{Dataset and Annotation}
\label{sec:dataset}
\subsection{Data Collection}
The dataset consists of 51 annotated reflective dialog sessions, drawn from a two-stage technology-mediated training study, illustrated here with a teacher training scenario as its concrete context~\cite{chehayeb2025effectscoregulationmodelmr}. In the first stage, participants took on a professional role, in this case a substitute teacher, in a simulated conflict scenario with virtual agents. In the second stage, a trained human interviewer guided each participant through a post-interaction interview, replaying critical moments of their recorded behavior and asking them to reflect on their experience and the motives behind it, in a setting explicitly framed as safe and non-judgmental. The present work targets self-compassion as it appears during this second, reflective stage. Session durations range from 600s to 4{,}956s (mean $\approx$3{,}489s). Audio and video are synchronized, and transcripts are derived via automatic speech recognition and aligned by timestamp across modalities.

\subsection{Annotation Scheme}
\begin{figure}[t]
\centering
\includegraphics[width=\linewidth]{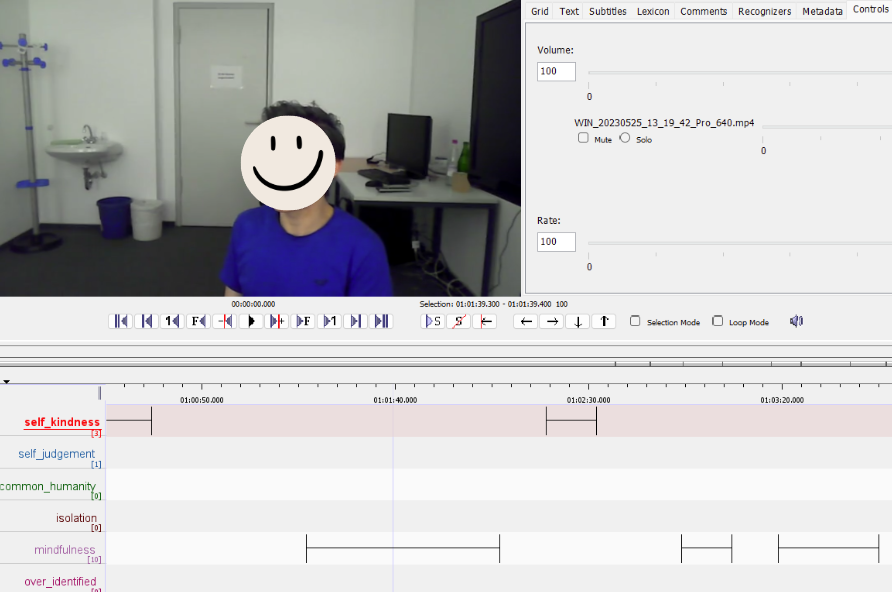}
\caption{Example ELAN annotation interface showing independent tiers for self-compassion components with permitted temporal overlap.}
\label{fig:elan}
\end{figure}

Each psychological construct was annotated as an independent ELAN~\cite{elan} tier with start and end timestamps. Temporal overlap between tiers was permitted: an interval labeled with one construct did not preclude simultaneous labeling with another, as shown in Figure~\ref{fig:elan}. To support annotation consistency, approximately 15\% of sessions were independently coded by a second annotator, with discrepancies discussed and resolved before finalizing the scheme; we did not compute a numeric agreement statistic such as Cohen's kappa.

\begin{table}[h]
\caption{Total annotated duration and video coverage per original construct.}
\label{tab:constructs}
\begin{tabular}{lcc}
\toprule
Construct & Total Duration (s) & \# Videos \\
\midrule
MF (Mindfulness) & 7136.6 & 51 \\
SK (Self-Kindness) & 3055.7 & 38 \\
SJ (Self-Judgment) & 650.3 & 16 \\
OI (Over-Identification) & 399.9 & 7 \\
CH (Common Humanity) & 121.5 & 5 \\
ISO (Isolation) & 15.0 & 1 \\
\bottomrule
\end{tabular}
\end{table}

An overlap and coverage audit, quantifying interval counts, total duration, per-session coverage, and pairwise temporal overlap per construct, confirmed the imbalance in Table~\ref{tab:constructs}: MF and SK account for the majority of annotated duration, SJ and OI show moderate but uneven distribution, and CH and ISO have limited coverage across sessions, owing to the fact that the interview targeted the activation of positive self-compassion sub-constructs. These findings directly motivated the consolidation described in Section~\ref{sec:problem}.

\subsection{Window-Based Supervision}
Annotated sessions are segmented into overlapping 4-second windows with a 1-second hop. A window is labeled positive for class $c$ if
\[
\frac{\text{duration}(w \cap a_c)}{\text{duration}(w)} \geq 0.2,
\]
where $a_c$ denotes an annotation interval of class $c$; for a 4-second window this corresponds to at least 0.8 seconds of overlap. This threshold reduces sensitivity to marginal boundary effects while retaining short but meaningful annotations. Segmentation across all 51 sessions produced 155{,}960 total windows, of which 10{,}658 (6.83\%) carry at least one positive label.

\subsection{Splits}
Splits are defined at the video level, not the window level, to prevent temporal leakage, using an 80/10/10 train/validation/test ratio with a fixed seed, stratified with respect to the original six annotated constructs prior to consolidation. This yields 24{,}249 training, 3{,}450 validation, and 4{,}275 test windows used for model fitting and selection; evaluation in Section~\ref{sec:results} is further restricted to the subset of test windows carrying at least one valid supervision label ($n=1{,}425$). All statistical fitting, standardization and imputation, is performed exclusively on the training split. The 24{,}249/3{,}450/4{,}275 windows reflect those falling within self-compassion construct segments rather than the full 155{,}960 windows generated across entire session recordings, which include stretches of the interview outside any annotated construct.
\section{Multimodal Preprocessing and Alignment}
\label{sec:pipeline}

Recordings arrive at heterogeneous resolutions, frame rates, and bitrates, so all sessions are first standardized, video re-encoded to a common resolution and codec, audio resampled to a consistent rate, before any cross-modal alignment is attempted. All modalities then share a common temporal reference derived from these standardized recordings. Alignment proceeds through speaker diarization, automatic speech recognition, and interval-based matching to ELAN annotations.

\begin{figure}[t]
\centering
\includegraphics[width=\linewidth]{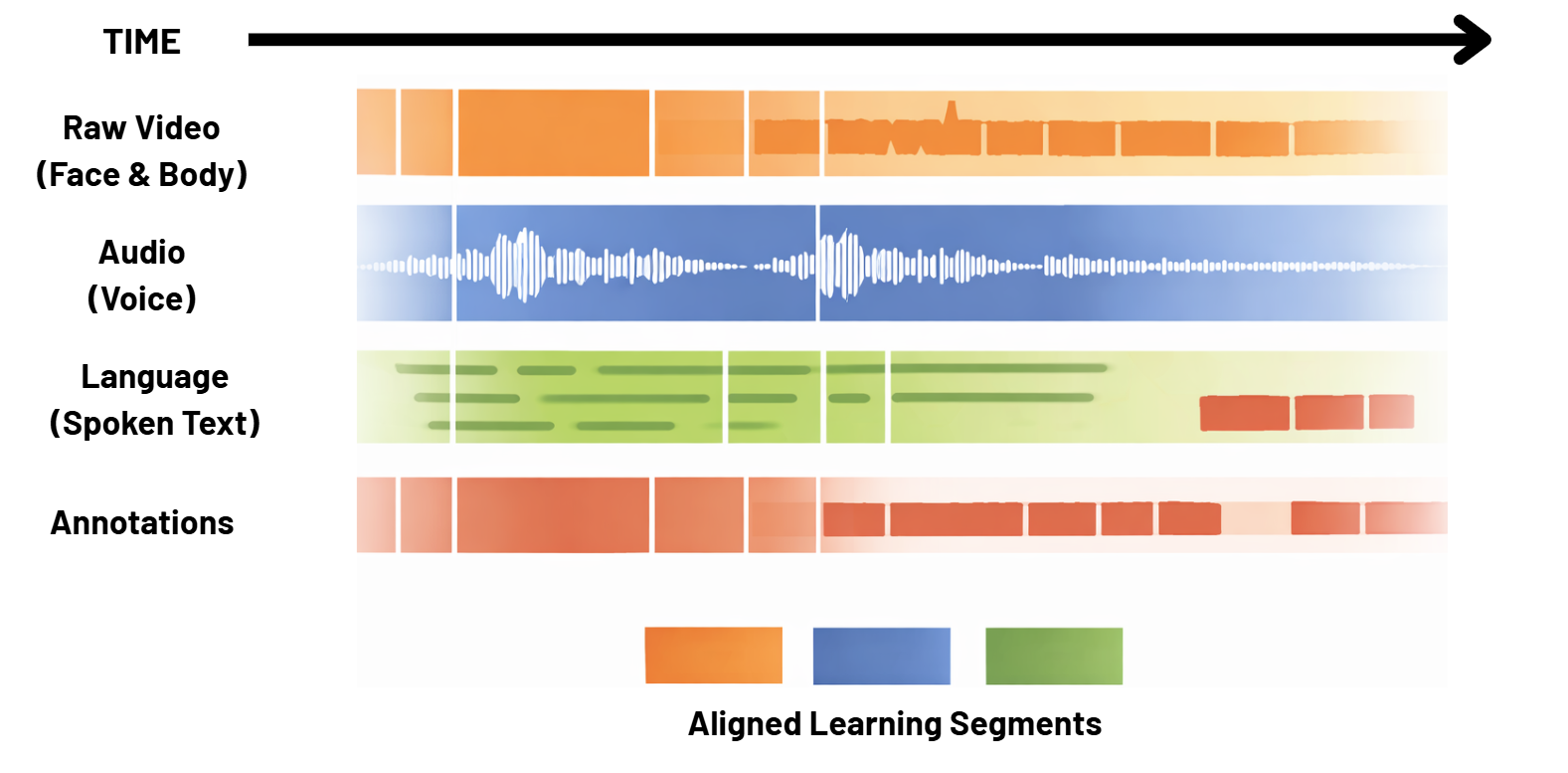}
\caption{Temporal alignment between raw video, audio, transcribed language, and ELAN annotations into aligned learning segments.}
\label{fig:alignment}
\end{figure}
Figure~\ref{fig:alignment} illustrates this alignment backbone across modalities.

\subsection{Video}
Facial blendshape coefficients and upper-body pose landmarks are extracted at 10 frames per second using MediaPipe~\cite{mediapipe2019}, yielding 40 time steps per 4-second window.

\begin{figure}[t]
\centering
\includegraphics[width=0.8\linewidth]{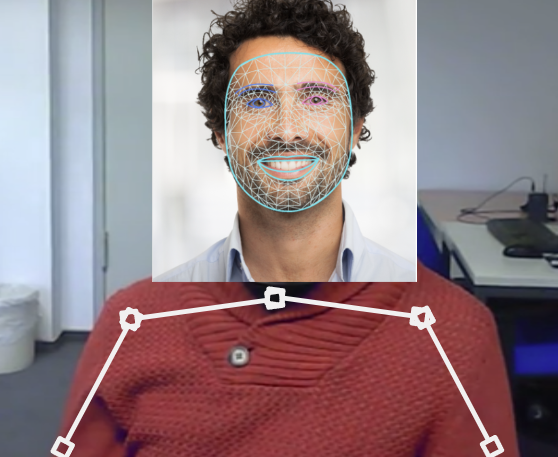}
\caption{Illustration of MediaPipe-based facial blendshape and upper-body pose landmark extraction.}
\label{fig:mediapipe}
\end{figure}
Figure~\ref{fig:mediapipe} illustrates this extraction process.

\subsection{Audio}
Speaker diarization is performed with \texttt{pyannote.audio}~\cite{pyannote2023}, and acoustic features are extracted with openSMILE using the eGeMAPSv02 configuration~\cite{opensmile2010,egemaps2016}, aligned to windows via temporal overlap and pooling.

\subsection{Text}
Automatic speech recognition is performed with Faster-Whisper (medium model, German)~\cite{whisper2023}. Text classification operates at the transcript segment level; segment-level predictions are subsequently projected onto the fixed 4-second windows for fusion and evaluation.
\section{Modeling and Fusion}
\label{sec:modeling}

Each modality is modeled independently under the same multi-label formulation, producing three independent sigmoid outputs for \{MF, SK, NEG\}, and evaluated using the window-level protocol of Section~\ref{sec:dataset} (with text projected from the segment level).

\subsection{Unimodal Models}
\textbf{Video:} A single-layer GRU~\cite{gru2014} with hidden size 128 processes the 40-step frame sequence within each window; the final hidden state feeds a fully connected layer with three sigmoid heads and a dropout of 0.2. The model is trained with AdamW (learning rate $5\times10^{-4}$, weight decay $2\times10^{-2}$) for up to 30 epochs with early stopping on validation loss, using per-class positive weighting in the binary cross-entropy loss to address class imbalance.

\textbf{Audio:} A logistic regression model (L2 penalty, \texttt{liblinear} solver, $C{=}1.0$), implemented as three independent one-vs-rest classifiers over standardized eGeMAPS functionals with per-class class-weighting, serves as the primary baseline; an exploratory MLP (hidden size 256) did not yield consistent gains and is not reported here in detail.

\textbf{Text:} A LLaMA-3.2 instruction model (3B parameters)~\cite{llama3} is adapted with QLoRA~\cite{qlora2023} (rank 16, alpha 32, 4-bit NF4 quantization) for parameter-efficient supervised fine-tuning at the transcript-segment level, using context-enhanced input (target segment plus preceding transcript context) over 3 epochs with an effective batch size of 4.

\subsection{Evaluation Protocol}
Each model produces independent class probabilities $\hat{p} \in [0,1]^3$ without cross-class normalization. Threshold-dependent metrics (precision, recall, F1) use a fixed decision threshold of 0.5; PR-AUC and ROC-AUC are computed directly from continuous probabilities. Macro-averaged metrics are the unweighted mean across classes; micro-averaged metrics aggregate true/false positives across classes before computing the metric. Model selection uses validation performance exclusively; no test-set information informs hyperparameter tuning, thresholding, or fusion weight selection.

\subsection{Late Fusion}
Given the limited number of annotated sessions and heterogeneous modality representations, we do not pursue joint end-to-end multimodal training. Instead, for modality set $M \subseteq \{\text{Video}, \text{Audio}, \text{Text}\}$ and class $c \in \{\text{MF}, \text{SK}, \text{NEG}\}$, each modality $m \in M$ produces a probability $p_{m,c}$, and the fused probability is
\[
\hat{p}_c = \sum_{m \in M} w_m\, p_{m,c}, \qquad w_m \geq 0,\ \sum_{m \in M} w_m = 1,
\]
with weights shared across classes. Weight configurations are selected by validation macro-F1 on a fixed grid and applied without modification to the held-out test split; no unimodal models are retrained during fusion.

\begin{figure}[t]
\centering
\includegraphics[width=\linewidth]{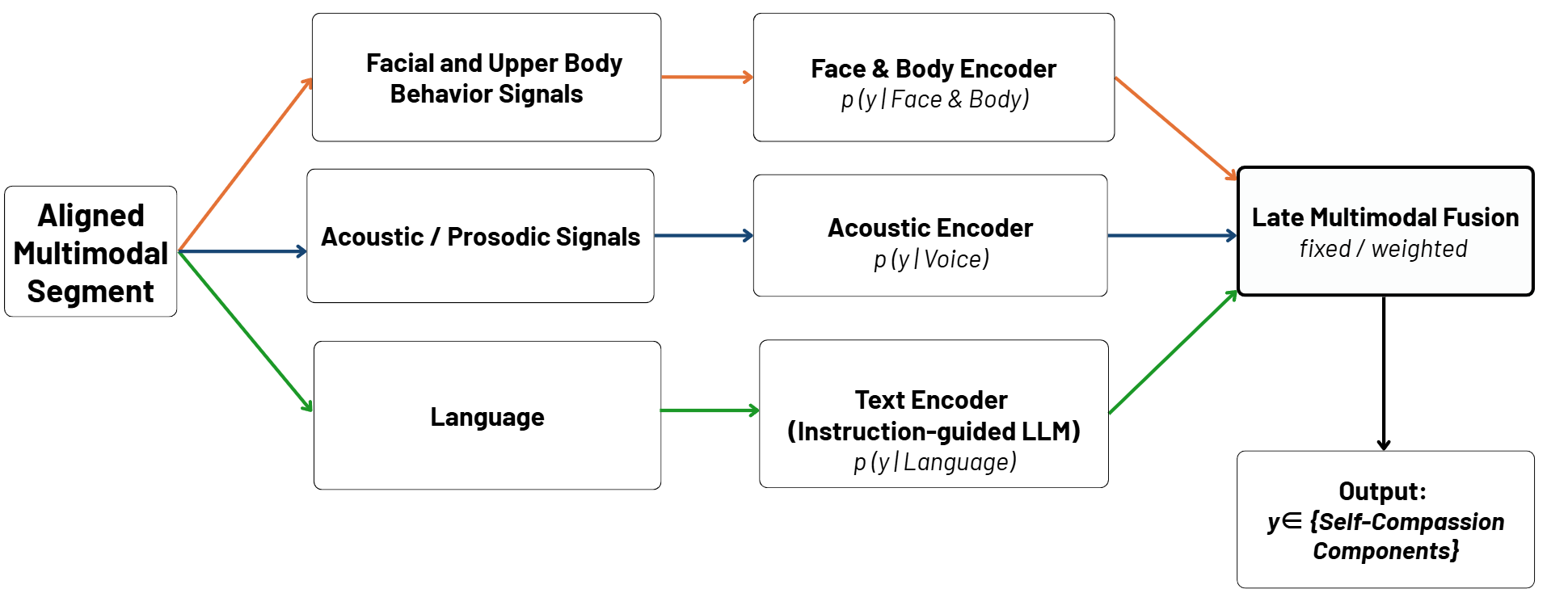}
\caption{Probability-level late fusion: independent unimodal encoders produce $p(y \mid \text{modality})$, combined via fixed or validation-selected weighting.}
\label{fig:fusion}
\end{figure}
Figure~\ref{fig:fusion} summarizes this architecture.
\section{Results}
\label{sec:results}
Evaluation is restricted to windows carrying at least one valid supervision label in the test split ($n=1{,}425$), with class support of 0.687 (MF), 0.124 (SK), and 0.051 (NEG). Threshold-dependent metrics use a fixed decision threshold of 0.5.

\begin{table}[h]
\caption{Window-level unimodal performance (test split). M-F1: macro-F1; $\mu$-F1: micro-F1.}
\label{tab:unimodal}
\small
\begin{tabular}{lcccc}
\toprule
Model & M-F1 & $\mu$-F1 & PR-AUC & ROC-AUC \\
\midrule
Video (GRU) & 0.331 & 0.610 & 0.30 & 0.77 \\
Audio (LogReg) & 0.252 & 0.379 & 0.15 & 0.47 \\
\bottomrule
\end{tabular}
\end{table}

\begin{table}[h]
\caption{Majority-class baseline (test split), derived from class prevalence.}
\label{tab:majority}
\begin{tabular}{lccc}
\toprule
Class & Prevalence & Majority Label & F1 \\
\midrule
MF & 0.687 & Positive & 0.814 \\
SK & 0.124 & Negative & 0.000 \\
NEG & 0.051 & Negative & 0.000 \\
\bottomrule
\end{tabular}
\end{table}
Table~\ref{tab:majority} reports the F1 achieved by always predicting the majority label per class, computed analytically from class prevalence. For SK and NEG, both unimodal models comfortably exceed this trivial baseline. For MF, however, the majority baseline (0.814) exceeds the strongest unimodal model's F1 (Video, 0.723), indicating that F1 alone is a misleading measure of learned discrimination on this class given the severe imbalance; PR-AUC and ROC-AUC (Table~\ref{tab:unimodal}) should be weighted more heavily than F1 when interpreting MF performance.

\begin{table}[h]
\caption{Per-class F1 (window-level, test split).}
\label{tab:perclass}
\begin{tabular}{lccc}
\toprule
Model & MF F1 & SK F1 & NEG F1 \\
\midrule
Video & 0.723 & 0.224 & 0.046 \\
Audio & 0.522 & 0.234 & 0.000 \\
\bottomrule
\end{tabular}
\end{table}
Audio fails to detect NEG entirely at the 0.5 threshold (F1 = 0.000), consistent with NEG's low support (0.051) and audio's comparatively weak overall discrimination (Table~\ref{tab:unimodal}); we return to this in Section~\ref{sec:discussion}.

Text is evaluated separately at the segment level, since its supervision unit differs from the window-level video and audio models. The fine-tuned model reaches a macro-F1 of 0.49 at the segment level (MF F1 0.58, SK F1 0.38, NEG F1 0.50), the highest macro-F1 among individual modalities, reflecting the more direct semantic access that linguistic content provides for explicitly verbalized reflective states.

\begin{table}[h]
\caption{Late fusion comparison (test split, window-level).}
\label{tab:fusion}
\begin{tabular}{lcc}
\toprule
Configuration & Macro-F1 & Micro-Recall \\
\midrule
Audio (unimodal) & 0.321 & 0.434 \\
Video (unimodal) & 0.331 & 0.606 \\
Text + Audio & 0.340 & 0.461 \\
Text + Video & 0.335 & 0.618 \\
Audio + Video & 0.345 & 0.633 \\
Text + Audio + Video & \textbf{0.350} & \textbf{0.645} \\
\bottomrule
\end{tabular}
\end{table}

\begin{table}[h]
\caption{Validation-selected fusion weights.}
\label{tab:weights}
\begin{tabular}{lccc}
\toprule
Configuration & $w_{\text{video}}$ & $w_{\text{audio}}$ & $w_{\text{text}}$ \\
\midrule
Video + Audio & 0.85 & 0.15 & -- \\
Video + Text & 0.85 & -- & 0.15 \\
Audio + Text & -- & 0.85 & 0.15 \\
Tri-modal & 0.33 & 0.33 & 0.33 \\
\bottomrule
\end{tabular}
\end{table}
Validation-selected weights favor video in every pairwise configuration that includes it ($w_{\text{video}} = 0.85$), while the tri-modal configuration is best served by uniform weighting ($w_m = 0.33$ for all $m$). The tri-modal fusion achieves the highest macro-F1 (0.350), an absolute improvement of 0.019 over the strongest window-level unimodal baseline (Video, 0.331), and also improves micro-recall over every unimodal or pairwise configuration.

\begin{figure}[t]
\centering
\includegraphics[width=\linewidth]{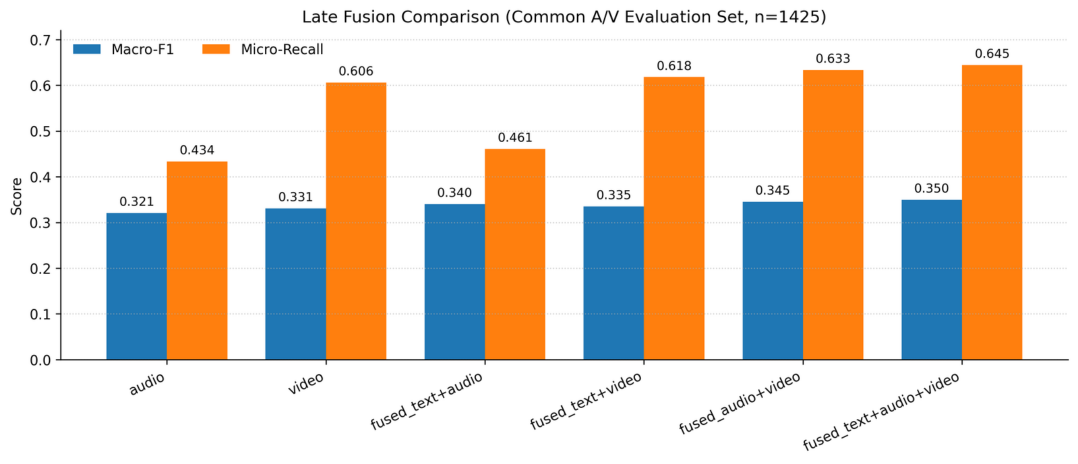}
\caption{Macro-F1 and micro-recall across unimodal and late fusion configurations (test split, window-level).}
\label{fig:fusioncomp}
\end{figure}
\section{Discussion}
\label{sec:discussion}

The results split cleanly along two lines: how well a class is supported in the annotation, and what kind of signal each modality actually has access to, behavioral, acoustic, or linguistic.

\subsection{Class-Specific Behavior}
\textbf{Mindfulness (MF)} has the highest support ratio (0.687 of evaluated windows) and the most stable learning behavior across modalities, consistent with the general pattern in modeling abstract, theory-driven constructs: better-supported classes with clearer behavioral or lexical correlates perform closer to well-studied affective categories, while sparser, more inferential classes lag behind~\cite{bailey2023,bailey2024}. That stability comes with a catch: MF's high prevalence puts the majority-class baseline at F1 0.814 (Table~\ref{tab:majority}), above every trained unimodal model's F1 on this class. So MF's strong F1 numbers owe more to class imbalance than to learned discrimination, and PR-AUC and ROC-AUC (Table~\ref{tab:unimodal}) are the better guide to what the models actually capture here. Text-based modeling performs consistently for MF, as reflective awareness and descriptive language provide accessible lexical cues; video also captures non-verbal correlates of reflective engagement within short windows.

\textbf{Self-Kindness (SK)} has substantially lower support, which increases sensitivity to modeling choices. Text modeling captures explicit supportive self-address when present, but the same sentiment is often expressed in different words across speakers, contributing to variability; video and audio capture indirect affective correlates but show limited standalone reliability. SK's low prevalence puts its majority-class baseline at zero, so unlike MF, every point of F1 the models earn here reflects real signal.

\textbf{Negative components (NEG)} aggregate self-judgment and over-identification and retain the lowest support ratio (0.051) despite this aggregation. Text modeling benefits from explicit evaluative language, but lexical ambiguity between reflective acknowledgment and evaluative judgment contributes to classification difficulty: a statement like "I didn't handle that well" can read as balanced acknowledgment of a mistake or as harsh self-judgment, depending on tone and context. Video and audio may capture tension or heightened affect, but such cues are indirect and overlap with other states. As with SK, NEG's majority baseline is zero, so the modest scores here are genuine, not an artifact of imbalance.

\subsection{Modality-Specific Strengths and Constraints}
Video captures frame-level facial activity and upper-body posture within 4-second windows via recurrent sequence modeling, which supports detection of short-term behavioral dynamics but limits modeling of gradual affective trajectories that unfold over longer spans. Audio, based on eGeMAPS functionals aggregated over diarization-defined segments, captures paralinguistic information that is less directly tied to construct semantics, and speaker variability introduces additional noise; the limited gap between logistic regression and a non-linear MLP suggests added model capacity does not substantially change discriminative capacity under current data conditions. Text is evaluated at the segment level rather than the window level, because a 4-second window often contains too little text, sometimes just a word or two, to carry reliable meaning on its own. This gives text the most direct semantic access to reflective and evaluative content, consistent with findings in adjacent domains that linguistic markers such as first-person pronoun use and self- versus other-focused language carry meaningful signal about self-relating and interpersonal stance in spoken interaction~\cite{ryu2023}. Its constraints are the flip side of this design: automatic transcription errors, and the need to project segment-level predictions onto windows for fusion. No single modality dominates across classes, and performance differences are class-dependent, reflecting how constructs are differentially expressed in reflective dialog.

\subsection{Interpretation of Fusion Behavior}
Incremental gains under pairwise and tri-modal fusion indicate partial complementarity: when one modality provides uncertain predictions, another provides supporting evidence. Validation-selected weights favor video in every pairwise configuration that includes it (Table~\ref{tab:weights}), so most of the discriminative signal at the window level is behavioral, with audio and text adding correction on top rather than contributing an equal share. The uniform tri-modal weighting fits this picture: once all three are available, none of them fully replaces the others, but the gain each adds beyond video is modest. Weight exploration on the validation split showed smooth performance variation, and test performance aligned with validation trends, suggesting fusion did not rely on unstable configurations. Probability-level late fusion avoids joint optimization across heterogeneous feature spaces, which, given the limited number of sessions and pronounced class imbalance, prioritizes stability and interpretability over the added capacity of joint multimodal architectures.
\section{Limitations and Future Work}
\label{sec:limitations}
Several of the limitations in this section follow directly from working with a newly collected, real-world corpus rather than an established benchmark, and are worth stating plainly rather than folding into the discussion above.
\begin{itemize}
\item \textbf{Effective sample size:} Sliding-window extraction increases the number of labeled instances (155{,}960 windows), but statistical independence is determined at the session level; the effective sample size is bounded by the 51 annotated sessions rather than the number of windows.
\item \textbf{Class imbalance and its source:} Two of the original six theoretical components, Isolation and Common Humanity, were excluded from modeling due to sparse coverage (Section~\ref{sec:problem}, Table~\ref{tab:constructs}). This imbalance is not purely a sampling artifact: the reflective sessions were designed to encourage constructive reflection and perspective-taking, which plausibly reduces the frequency of isolating or negatively valenced expression within this structured setting. The same imbalance also means F1 is a weaker measure of model quality on the best-supported class than on the others, as the majority-class baseline in Section~\ref{sec:results} shows.
\item \textbf{Single-corpus evaluation:} All results are corpus-specific to one structured, interviewer-guided training environment; generalization to spontaneous reflection or other educational or alternative settings has not been assessed.
\item \textbf{Fixed temporal granularity:} The 4-second, 1-second-hop windowing imposes a uniform discretization on reflective processes of inherently variable duration and may not align with construct boundaries.
\item \textbf{ASR and diarization noise:} Text-based modeling relies on automatic speech recognition without manual correction, and speaker diarization is similarly imperfect; both introduce an upper bound on achievable text performance and add alignment noise that is applied consistently but not eliminated.
\item \textbf{Evaluation design:} Evaluation is descriptive (F1, PR-AUC, ROC-AUC); no statistical significance testing or split-sensitivity analysis was performed. Probability calibration was also not performed, as the focus here was relative discriminative performance rather than deployment-ready probability estimates.
\item \textbf{Fusion capacity:} Probability-level late fusion with shared, static weights is deliberately low-capacity; it does not model cross-modal feature interactions and treats classes symmetrically in weighting.
\end{itemize}
Future work includes cross-corpus and cross-domain validation, more expressive and class-aware fusion strategies, probability calibration and uncertainty estimation, and a closer analysis of what fine-tuned language models learn about theory-grounded linguistic markers of these constructs relative to their pretrained, non-fine-tuned counterparts, which we are pursuing as an extension of this work.
\section{Data and Code Availability}
\label{sec:availability}
Code for preprocessing, feature extraction, model training, and fusion will be released publicly upon acceptance. The dataset contains identifiable video and audio of human participants collected under a consent protocol that did not anticipate public redistribution (Section~\ref{sec:ethics}); data access beyond what is reported here is handled on a case-by-case basis and is not guaranteed. Sections~\ref{sec:dataset} and~\ref{sec:pipeline} report the annotation protocol, window construction, and feature extraction pipeline in sufficient detail to support replication on comparably collected data.
\section{Ethical Considerations}
\label{sec:ethics}
The sessions underlying this dataset involve recorded facial video, audio, and speech from human participants engaging with emotionally sensitive material (perceived mistakes, self-evaluation, professional difficulty). Data collection followed institutional ethics review and informed consent procedures, including participants' right to withdraw and to have recordings excluded from analysis. All modeling in this paper operates on de-identified, timestamp-aligned features rather than raw video or audio, and no identifying information is reported at the individual level. Because the underlying constructs concern psychological well-being, we treat this as a modeling and measurement study rather than a diagnostic or evaluative tool, and we do not claim that model predictions are suitable for individual-level clinical or pedagogical judgments about a specific participant. Data access and sharing constraints arising from this consent scope are discussed in Section~\ref{sec:availability}. The right panel of the teaser figure (Figure~\ref{fig:teaser}) is an AI-generated illustration and does not depict a real individual.
\section{Conclusion}
\label{sec:conclusion}
This paper set out to see whether a theory-driven, higher-order construct like self-compassion could be consolidated into a workable supervision scheme, aligned across video, audio, and text, and detected with a small, imbalanced corpus. It can, though not evenly: the six-component theory reduces to a three-class scheme without losing its grounding, and a shared preprocessing pipeline puts the three modalities on the same timeline despite their different sampling rates and noise, but how well each class is actually detected still tracks how much annotated support it had to begin with, and for the best-supported class, a majority-class baseline is a real competitor to the trained models. Fusion adds a small but real improvement over the best single modality.

The gains here are modest, and that is mostly a function of scale, 51 sessions is not much data for three modalities and imbalanced classes. Self-compassion was one way into this problem; the same approach should carry over to other constructs that are theory-grounded and hard to observe directly. More data, richer fusion, and calibrated evaluation are the obvious next steps, both to strengthen self-compassion detection itself and to test how well this operationalization and modeling approach generalizes to other higher-order constructs and application settings that share the same core difficulty: real, theory-grounded, and only indirectly observable.

\begin{acks}
This work was supported by the German Federal Ministry of Research, Technology and Space (BMFTR) under grant 16KIS2632K (Project MARSS).
\end{acks}
\bibliographystyle{ACM-Reference-Format}
\bibliography{references}

\end{document}